# Robust Matrix Completion State Estimation in Distribution Systems

Bo Liu, Hongyu Wu
Dept. Elec. & Comput. Engineering
Kansas State University
Manhattan, United States

Yingchen Zhang, Rui Yang, Andrey Bernstein
Power Systems Engineering Center
National Renewable Energy Laboratory
Golden, United States

*Abstract*—Due to the insufficient measurements in the distribution system state estimation (DSSE), full observability and redundant measurements are difficult to achieve without using the pseudo measurements. The matrix completion state estimation (MCSE) combines the matrix completion and power system model to estimate voltage by exploring the low-rank characteristics of the matrix. This paper proposes a robust matrix completion state estimation (RMCSE) to estimate the voltage in a distribution system under a low-observability condition. Tradition state estimation weighted least squares (WLS) method requires full observability to calculate the states and needs redundant measurements to proceed a bad data detection. The proposed method improves the robustness of the MCSE to bad data by minimizing the rank of the matrix and measurements residual with different weights. It can estimate the system state in a low-observability system and has robust estimates without the bad data detection process in the face of multiple bad data. The method is numerically evaluated on the IEEE 33-node radial distribution system. The estimation performance and robustness of RMCSE are compared with the WLS with the largest normalized residual bad data identification (WLS-LNR), and the MCSE.

## I. Introduction

With the integration of distributed energy resources (DERs), aggregated demand response and electric vehicle (EV) charging in the distribution system, system operators need real-time monitoring to maintain the system reliability and efficiency in the face of more variable loads. The system state estimation (SE) is an essential tool for online monitoring and analysis in the transmission system, where measurement redundancy ensures the system observability and bad data processing. Even though the deployment of various recent technologies such as advanced metering infrastructure (AMI) phasor measurements units (PMUs), intelligent electronic devices and smart inverters of DERs, have improved the network observability, the distribution system is generally underdetermined with poor observability and easily becomes unobservable due to the communication failure and delay [1]. While pseudo-measurements based on the history of the distribution system are generally used to improve the system observability, pseudo-measurement errors significantly affect the estimation accuracy. This is the reason why only a limited number of utilities have implemented the distribution system state estimation (DSSE) [2-3].

Weighted least squares (WLS) is a conventional method in the DSSE including bus-voltage based methods and branch-current based methods depending on the selection of the state variables [4]. Although the WLS-based methods are fast, simple and broadly used, they require the full network observability and are also sensitive to the bad data. Data-driven approaches and machine-learning techniques are employed in the DSSE. A matrix completion state estimation (MCSE) was proposed in [5], [6] which used a matrix completion algorithm augmented with power-flow constraints to estimate the voltage in a low-observability system. However, the MCSE is also sensitive to bad data.

Bad data detection (BDD) is an integral function for the SE to detect, identify and correct measurement errors. The WLS combined with the largest normalized residual bad data identification (WLS-LNR) is widely used in power system control center [7]. The detectability of bad data depends heavily on the measurement configuration and redundancy [4]. Through a residual analysis, erroneous redundant measurements are identifiable, while inaccurate critical measurements that negatively affect the estimation state are undetectable. No critical measurements should exist in a well-designed measurement system [8]. However, the lack of real-time measurements in the distribution system results in low measurement redundancy, increased critical measurements, and even low network observability, which creates additional obstacles to the BDD. The undetectable erroneous critical measurement can deviate the estimated state from the actual value in the WLS based method.

A robust state estimation algorithm without post-SE bad data processing can be used in the DSSE to reduce the influence of bad data. Different from the WLS, robust estimators use different objective functions to improve the SE robustness to bad data, such as Least Median of Square, Least Trimmed Squares and Least Absolute Value [9-10]. Robust estimators generally reduce the weight of the erroneous measurement identified by a high residual in the estimation process to suppress the impact of bad data on the solution. In [11], the machine learning and state estimation were combined within a closed-loop scheme, in which the nodal load estimates from the state estimator is used by the machine learning function as a feedback to improve the accuracy.

By extending our prior work [5], [6], this paper tries to investigate and enhance the robustness of the MCSE. The basic idea is to minimize the weighted sum of matrix rank and measurements residual, instead of imposing a fixed threshold to the measurement residual. The proposed method is formulated as a convex optimization problem and solved by a semidefinite programming (SDP) solver. Estimation mean absolute percentage error (MAPE) of RMCSE is compared with those of the WLS and the MCSE on the IEEE 33-node distribution system with various system observability levels. Also, the robustness of RMCSE is shown in contrast to the WLS-LNR in the case study. The rest of this paper is organized as follows. The formulation for the RMCSE is presented in Section II. The case study is conducted in Section III. The paper is summarized and concluded in Section IV.

## II. PROBLEM FORMULATION

The background of matrix completion is presented and the formulation for RMCSE is proposed in this section.

### A. Matrix Completion

Matrix completion takes advantage of the low-rank property of the matrix to estimate the missing elements using the known elements in the matrix. Let $M \in \mathbb{R}^{n_1 * n_2}$ be the matrix we want to recover. In the $M$ matrix, only a sampled set of entries $M_{ij}, (i,j) \in \Omega$ are available, and the rest entries are unknown, where $\Omega$ is a subset of the complete set of M. A sampling operator $P_\Omega : \mathbb{R}^{n_1 * n_2} \rightarrow \mathbb{R}^{n_1 * n_2}$ is generally used to represent the available information via $P_\Omega(M)$:

$$[P_\Omega(X)]_{ij} = \begin{cases} X_{ij}, & (i,j) \in \Omega \\ 0, & otherwise \end{cases} \quad (1)$$

Assume there is a low-rank matrix $X \in \mathbb{R}^{n_1 * n_2}$ which is consistent with the observed entries in $M$. Then, the rank minimization problem can be formulated to recover the unknown entries in $M$ as follows:

$$\begin{aligned} \min \quad & rank(X) \\ s.t. \quad & P_\Omega(X) = P_\Omega(M) \end{aligned} \quad (2)$$

This problem is NP-hard and its solution algorithms are doubly exponential. The nuclear norm can be used in the objective function to formulate the problem as a convex problem according to the convex relaxation [5]. Furthermore, equality constraints are relaxed using the Frobenius norm to suppress the influence of bad data. The matrix completion problem is formulated as follows:

$$\begin{aligned} \min \quad & \|X\|_* \\ s.t. \quad & \|P_\Omega(X) - P_\Omega(M)\|_F \leq \delta \end{aligned} \quad (3)$$

where the nuclear norm $\|X\|_* = \sum_{i=1}^{\min(n_1, n_2)} \sigma_i(X)$ is the sum of the matrix singular value, and the Frobenius norm is defined as:

$$\|P_\Omega(X) - P_\Omega(M)\|_F = \sqrt{\sum_{i=1}^{n_1}\sum_{j=1}^{n_2}\left|[P_\Omega(X)]_{ij} - [P_\Omega(M)]_{ij}\right|^2} \quad (4)$$

### B. Robust Matrix Completion State Estimation

A robust state estimation based on matrix completion is proposed to estimate the voltage in a distribution system with a low-observability. This approach combines the matrix completion and the power system knowledge by integrating all the system information into a system state-measurement matrix. The distribution system model and Ohm's Law are introduced and formulated as constraints to provide more insights into missing and observed entries.

In a system state-measurement matrix, each row integrates the information of one feeder, and each column represents one measurement variable. In greater detail, the columns include the real and reactive voltage of the feeder's two end buses, the voltage magnitude of the two buses, the active and reactive power injection of two buses, and the real and reactive part of line current. For each feeder $(i,j)$, its corresponding row in the matrix is defined as:

$$[\text{Re}(V_i), \text{Im}(V_i), |V_i|, P_{in}^i, Q_{in}^i, \text{Re}(V_j), \text{Im}(V_j), |V_j|, P_{in}^j, Q_{in}^j, \\ \text{Re}(I_{ij}), \text{Im}(I_{ij})] \quad (5)$$

In a distribution system with $n$ nodes and $m$ feeders, system state-measurement matrix $M \in \mathbb{R}^{m*12}$ is composed by the system states and measurements. All measurements obtained from the sensors are observed entries in the matrix, while the rest of the measurements and system states can be recovered by the matrix completion. The RMCSE is formulated as a convex optimization problem:

$$\min \quad \|X\|_* + w_1 \|P_\Omega(X) - P_\Omega(M)\|_F + w_2 \sum_{(i,j)\in\zeta} \varepsilon_{ij} + w_3 \gamma + w_4 \alpha$$

$$\begin{aligned} s.t. \quad & \left|I_{ij} - (V_i - V_j)Y_{ij}\right| \leq \varepsilon_{ij}, \quad \forall (i,j) \in \zeta \\ & \|v - Dx - w\|_\infty \leq \gamma \\ & \||v| - Kx - |w|\|_\infty \leq \alpha, \\ & \varepsilon_{ij} \geq 0, \quad \forall (i,j) \in \zeta \end{aligned} \quad (6)$$

where $w_1, w_2, w_3$ and $w_4$ are the weight parameters, and $w_1 = 2$, $w_2 = w_3 = w_4 = 200$ in our case study; $\zeta$ is the set of distribution feeders; $Y_{ij}$ is the admittance of the feeder $(i,j)$; $v \in \mathbb{C}^n$ is the bus voltage phasor vector, $|v| \in \mathbb{R}^n$ is the bus voltage magnitude vector, $x \in \mathbb{R}^{2n}$ is a vector of the active and reactive power injection of all buses; $D \in \mathbb{C}^{n \times 2n}$, $K \in \mathbb{R}^{n \times 2n}$ and $w \in \mathbb{C}^n$ can be calculated [12].

The first constraint in (6) is the Ohm's Law of each feeder, which set up the relationship between the real and reactive part of voltage and current in the matrix. Instead of using linear equality constraints for the Ohm's Law, the constraints are relaxed by a tolerance $\varepsilon$ due to the measurement noise. The second and third constraints in (6) are the linear approximation of voltage phasor and voltage magnitude in the distribution system, respectively [12]. The nodal net power injection is used to approximate the power-flow solution based on a fixed-point linearization of the AC power-flow equation. Similarly, the constraints are relaxed by tolerance because the linear approximation itself has estimation errors, and the power injections used in the linear approximation have measurement error.

The proposed RMCSE (6) differs from the matrix completion (3) in the following two ways. First, unlike purely exploiting the low-rank property in (3), the RMCSE introduces constraints of the distribution system model and Ohm's Law, leading to the better recovery of the missing entries. Secondly, the Frobenius norm of the measurements is formulated in the objective function rather than in the constraints. If the Frobenius norm of the measurements is imposed as hard constraints in (3), one needs to carefully select the tolerance value $\delta$. An improper selection of $\delta$ will deviate the estimated states from the actual ones in the case of one or more bad data in the measurements. Therefore, formulating the Frobenius norm in the objective function can avoid the tolerance selection issue and increase the algorithmic robustness to the bad data.

## III. CASE STUDY

The proposed RMCSE is validated on the IEEE 33-node radial distribution system. The MatPower 6.0 is used to generate the measurements for the system. The measurement noise is the Gaussian distribution with a standard deviation of 1% of the actual value. The proposed RMCSE is modeled by the CVXPY [13], and the problem is solved by the SDP solver.

To reflect the number of measurements in the system, we define an indicator, namely fraction of the available data (FAD), as the ratio of the number of measurements used in the state estimation over the total number of all possible measurement in the system. In the 33-node system, the total number of measurements is 165, including 2 measurements of reference bus voltage phasor in rectangle coordinate, 33 voltage magnitude, 66 active and reactive power injection measurements and 64 measurements of line current in rectangle coordinate. For a given FAD, the measurements used in the SE are randomly selected from the 165 measurements, except for the voltage phasor at the reference bus.

### A. The performance of the RMCSE

In this section, the performance of the RMCSE is compared with that of the MCSE and the WLS in the following three systems: 1) an observable system with sufficient redundant measurements, 2) a low-observability system with several redundant measurements, and 3) an unobservable system with few measurements.

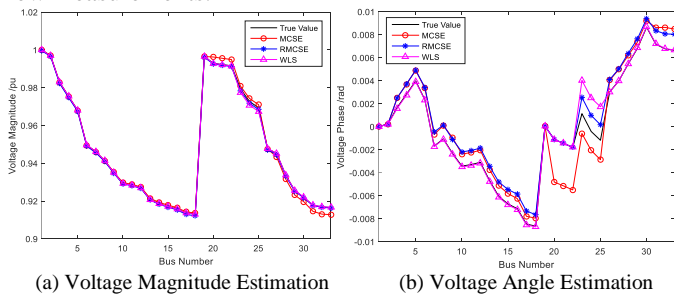

(a) Voltage Magnitude Estimation     (b) Voltage Angle Estimation

Fig. 1 The comparison among the RMCSE, MCSE and WLS (FAD=0.7).

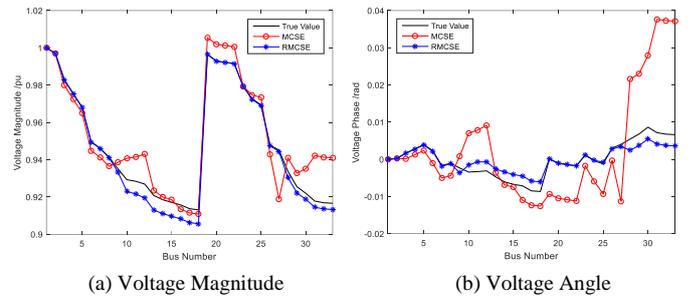

(a) Voltage Magnitude     (b) Voltage Angle

Fig. 2 The comparison of the RMCSE and MCSE (FAD=0.5).

When the FAD is 0.7, the 117 measurements are randomly selected. The redundancy factor is 1.75 which ensures an observable system and sufficient redundant measurements. The comparison of the RMCSE, MCSE, and WLS is shown in Fig. 1. All of these three methods can accurately estimate the voltage magnitude, but the estimated voltage angles in the RMSE and the MCSE has some slight offset in several buses. When the number of measurement reduce to 84 (0.5 FAD), the redundancy factor of the system becomes 1.24. The system becomes unobservable. The rank of the Jacobian matrix is 63 and there are 3 unobservable states in the system. The estimation performances of the RMCSE and the MCSE are shown in Fig. 2. It is seen that the RMCSE and the MCSE both deviate from the true value, and the RMCSE is better than the MCSE in the voltage angle estimation.

When the FAD decreases to 0.3, only 47 measurements can be used to estimate 66 system states. The redundancy factor is 0.74. The rank of the Jacobian matrix is 47 and there are 19 unobservable states in the system. The estimation results of the MCSE and the RMCSE are shown in Fig. 3. It is seen that the RMCSE outperforms the MCSE in the voltage angle estimation, whereas it underperforms the MCSE in the voltage magnitude estimation.

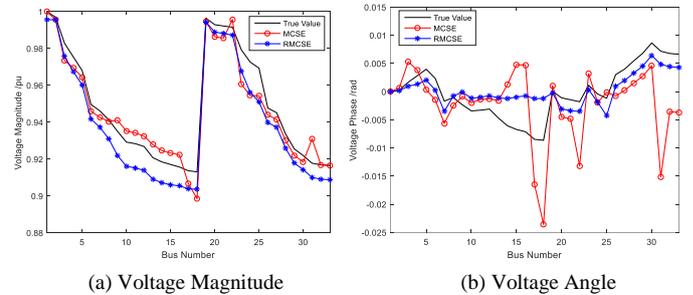

(a) Voltage Magnitude     (b) Voltage Angle

Fig. 3 The comparison between the RMCSE and the MCSE (FAD=0.3).

To illustrate the performance of these methods under different FADs, 30 cases with different measurement sets are generated for each FAD. In each case, measurements are randomly selected, and the MAPE of estimation result in each method is calculated. The mean MAPE of the 30 cases under different FADs is calculated and shown in Fig. 4. Note that the performance of the WLS is only shown in observable systems (FAD>0.5). It is seen that the MAPEs of these methods decrease with the increase of system observability. The RMCSE has the best voltage angle estimation performance, and its MAPE of voltage magnitude is the best among all three methods when the FAD is between 0.6-0.9. While the MCSE performs the best in voltage magnitude estimation in the low FAD, its angle estimation is far from accurate.

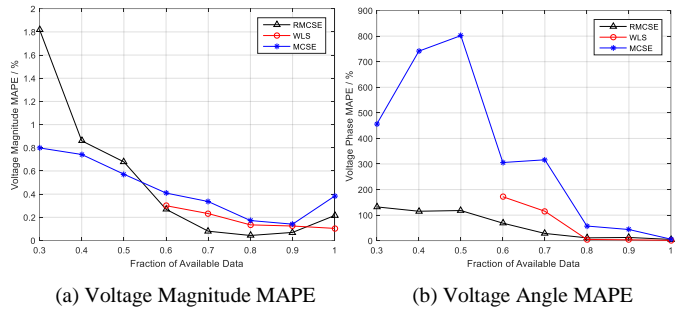

(a) Voltage Magnitude MAPE        (b) Voltage Angle MAPE

Fig. 4 The influence of FAD on the MAPE of voltage estimation.

*B. The robustness of the RMCSE*

To investigate the robustness of the RMCSE, the estimation results of the WLS and the RMCSE are compared when there exist bad data in the measurements. Assume that the active power injection of Node 18 is an erroneous measurement, and its value becomes two times of the actual value in a system with 0.7 FAD.

In Fig. 5, the influence of bad data on the WLS is shown by comparing the estimated states with and without bad data. It is seen in Fig. 5(a) that bad data causes all estimated voltage magnitude larger than their actual value. This is due to Bus 18 which is the end node of the main feeder and its active power injection can influence the voltage magnitude of all buses. To facilitate the comparison, all measurements and the bad data for the WLS and RMCSE are the same. The impact of bad data on the RMCSE is shown in Fig. 6. It is observed that most of the states remain unchanged except for slight changes in the voltage phase of Buses 16 and 17, as shown in Fig. 6(b). The implication of the results is twofold. First, the WLS is sensitive to the bad data. Therefore, the bad data ought to be identified and deleted before the application of the WLS. Second, the bad data has a very limited influence on the RMCSE. The RMCSE can obtain accurate system states without identification and removal of the bad data. The characteristics of RMCSE shows its robustness to the bad data.

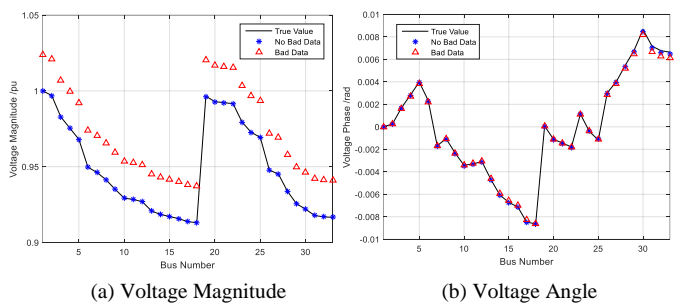

(a) Voltage Magnitude        (b) Voltage Angle

Fig. 5 Estimation by the WLS with and without bad data.

Since the WLS is sensitive to the bad data, we further compare the RMCSE with the WLS-LNR which can identify and delete the bad data based on the residual analysis and is therefore robust to the bad data. Here, we investigate the impact of multiple bad data on the SE methods by creating 11 tests with a bad data percentage (out of the total measurements) starting from 0% to 10% with an increment of 1%. For each percentage, 30 cases are generated with the same measurements but different bad data. The location and value of the bad data in these cases are randomly generated. Assume the noise distribution of normal measurements is Gaussian with a standard deviation of 1% of the actual value, and the standard deviation of bad measurements is 100% of the actual value, which is 100 times that of the normal measurements.

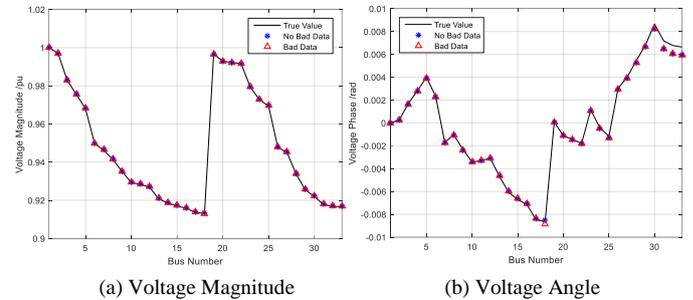

(a) Voltage Magnitude        (b) Voltage Angle

Fig. 6 Estimation by the RMCSE with and without bad data.

When the FAD is 0.7, the MAPEs of the voltage magnitude and angle versus the bad data percentage are compared in Fig. 7. In Figs. 7(a) and 7(b), it shows that the MAPE of the WLS dramatically increases with the bad data percentage. In the WLS, bad data deviates the estimated state from the actual states seriously. The performance of WLS-LNR is much better than the WLS. It shows that the voltage magnitude MAPE slightly increases with the bad data percentage, and angle MAPE increases with bad data percentage. In the WLS-LNR, the bad data has a less negative impact on the estimation results, because all of the bad data are identified and removed.

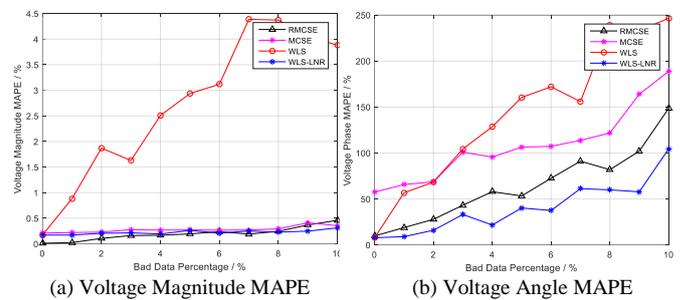

(a) Voltage Magnitude MAPE        (b) Voltage Angle MAPE

Fig. 7 The comparison of voltage MAPE with multiple bad data (FAD=0.7).

Compared with the WLS, the RMCSE leads to better MAPEs of the voltage magnitude and angle. This can be explained by comparing their objective functions. The WLS solely minimizes the measurement residual, whereas the RMCSE minimizes the weighted sum of the matrix rank and measurement residual, and a small weight is usually assigned to the measurement residual in (6). When comparing the RMCSE with the WLS-LNR, the RMCSE leads to better MAPEs of the voltage magnitude but worse MAPEs of the voltage angle. It is observed in Fig. 7(b), when the FAD is 0.7 (the system has redundant measurements), the WLS-LNR is the best SE method since it has enough measurements to yield a relatively accurate estimation after deleting the bad data.

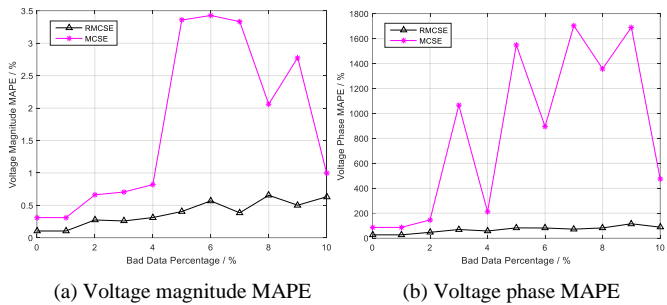

(a) Voltage magnitude MAPE   (b) Voltage phase MAPE

Fig. 8 The comparison of voltage MAPE with multiple bad data (FAD=0.5).

When the FAD is 0.5, i.e., the system has a low-observability, the performance of those approaches with the multiple bad data are shown in Fig. 8. The rank of the Jacobian matrix is 62 and there are 4 unobservable states in the system. It is seen in Fig. 8 that the RMCSE has a better performance in both voltage magnitude and phase MAPE than the MCSE.

The comparison under 0.3 FAD is shown in Fig. 9. When FAD is 0.3, 51 measurements are used to estimate 66 states. Compared to the voltage MAPE of RMCSE and MCSE at FAD=0.5, voltage MAPE of RMCSE and MCSE both increase at FDA=0.3, because more measurements become critical measurements at FDA=0.3 and the bad data in critical measurements will have a bad impact on the estimation. The RMCSE has a better performance in both voltage magnitude and phase MAPE than the MCSE. The results demonstrate the robustness of MCSE in the low-observability system.

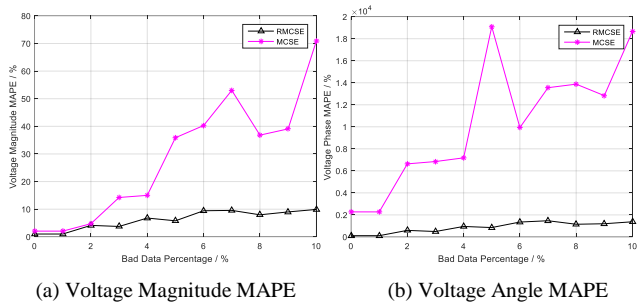

(a) Voltage Magnitude MAPE   (b) Voltage Angle MAPE

Fig. 9 The comparison of voltage MAPE with multiple bad data (FAD=0.3).

## IV. CONCLUSIONS

This paper proposes a robust matrix completion state estimation in distribution systems under low-observability. The proposed approach is formulated as a convex optimization problem considering the bad data influence. The case study compares the performances of RMCSE, MCSE, and WLS at the different levels of system observability. The results show that the RMCSE achieves more accurate estimation than the WLS in observable systems. When the system has multiple bad data, the proposed RMCSE has similar performance with WLS-LNR in observable system, and outperforms the MCSE in the estimation of voltage magnitude and angle in low-observability systems. The proposed method yields a robust SE without the need for post-SE bad data processing. PMU data and other useful measurements can also be seamlessly integrated into the proposed approach. In the future work, we will extend the proposed RMCSE to three-phase unbalanced distribution systems. The performance of different SE approaches and their robustness to the bad data will be thoroughly investigated in low-observable distribution systems.

ACKNOWLEDGMENT

Funding provided by U.S. Department of Energy Office of Energy Efficiency and Renewable Energy Solar Energy Technologies Office. The views expressed in the article do not necessarily represent the views of the DOE or the U.S. Government. The U.S. Government retains and the publisher, by accepting the article for publication, acknowledges that the U.S. Government retains a nonexclusive, paid-up, irrevocable, worldwide license to publish or reproduce the published form of this work, or allow others to do so, for U.S. Government purposes.